# Decoding Musical Origins: Distinguishing Human and AI Composers

An Analysis of Technological Fingerprints in YNote-Encoded Music

CHENG-YANG TSAI*

Dept. of Computer Science & Engineering, Yuan Ze University

s1111505@mail.yzu.edu.tw

TZU-WEI HUANG*

Dept. of Computer Science & Engineering, Yuan Ze University

s1113315@mail.yzu.edu.tw

SHAO-YU WEI

Dept. of Computer Science & Engineering, Yuan Ze University

s1123332@mail.yzu.edu.tw

GUAN-WEI CHEN

Dept. of Computer Science & Engineering, Yuan Ze University

s1111446@mail.yzu.edu.tw

HUNG-YING CHU

Dept. of Computer Science & Engineering, Yuan Ze University

s1111521@mail.yzu.edu.tw

YU-CHENG LIN*

Dept. of Computer Science & Engineering, Yuan Ze University

linyu@saturn.yzu.edu.tw

With the rapid advancement of Large Language Models (LLMs), AI-driven music generation has become a vibrant and fruitful area of research. However, the representation of musical data remains a significant challenge. To address this, a novel, machine-learning-friendly music notation system, YNote, was developed. This study leverages YNote to train an effective classification model capable of distinguishing whether a piece of music was composed by a human (Native), a rule-based algorithm (Algorithm Generated), or an LLM (LLM Generated). We frame this as a text classification problem, applying the Term Frequency-Inverse Document Frequency (TF-IDF) algorithm to extract structural features from YNote sequences and using the Synthetic Minority Over-sampling Technique (SMOTE) to address data imbalance. The resulting model achieves an accuracy of 98.25%, successfully demonstrating that YNote retains sufficient stylistic information for analysis. More importantly, the model can identify the unique " technological fingerprints " left by different AI generation techniques, providing a powerful tool for tracing the origins of AI-generated content.

**CCS CONCEPTS** • Applied computing → Sound and music computing • Computing methodologies → Machine learning • Information systems → Information retrieval

**Additional Keywords and Phrases:** AI Music Generation, YNote, TF-IDF, Source Classification

## 1 INTRODUCTION

In recent years, the rapid development of Large Language Models (LLMs) has turned artificial intelligence music generation into a vibrant and fruitful research field [1]. By fine-tuning pre-trained models on task-specific datasets, high-quality content can be generated across various domains. In music, studies have successfully used formats like ABC notation to interact with GPT-2 for music generation, achieving remarkable results [2]. However, the representation of musical data has always been a major challenge in this field. Traditional formats such as MIDI, MusicXML, or ABC notation have complex and variable structures, making them less than ideal as input for LLMs designed to process text. This complexity also increases the difficulty and uncertainty of model fine-tuning.

To overcome this data representation bottleneck, Lu et al. proposed YNote, a new music notation system designed for machine learning [3]. With its minimalist design, YNote uses a fixed four-character string to represent a note, including its pitch and duration. The high consistency and conciseness of this format show great potential when used as input for machine learning models. The core advantage of YNote lies in its high readability for machine learning models. Through its minimalist design, a complex piece of music is effectively translated into a structured "text" for a computer. This transformation greatly simplifies the data processing workflow and allows researchers to apply mature text analysis techniques for more diverse musical operations.

The main purpose of this study is to build upon this foundation by training an efficient classification model to accurately predict whether a piece of music encoded in YNote is of human origin (native) or AI-generated. Although YNote has proven its effectiveness in music generation tasks, a deeper question arises: does this simplified format still retain enough stylistic information to be used for music analysis and classification? When music from vastly different sources is converted into YNote sequences, can we still identify their unique "technological fingerprints"?

This research aims to fill this gap. We propose an innovative framework that treats YNote music sequences as a special kind of language text and introduces mature techniques from the field of Natural Language Processing (NLP) to solve this classification problem. Specifically, this study utilizes the Term Frequency-Inverse Document Frequency (TF-IDF) algorithm [4,5,6] to extract structural features from the music and employs the Synthetic Minority Over-sampling Technique (SMOTE) [7,8] to address the challenge of data imbalance in the real world. The results of this study not only validate the potential of YNote in music content analysis but also successfully establish a high-precision model that can not only distinguish between human and AI creations but also identify the unique "technological fingerprints" left by different AI technologies (rule-based algorithms vs. data-driven LLMs).

## 2 PROBLEM DESCRIPTION

As AI music generation technology matures, a critical question emerges: can we accurately determine whether a piece of music was created by a human or by artificial intelligence? Furthermore, when different AI technologies,such as rule-based algorithms and data-driven large language models,can both generate music, can we trace the technological origin from the music itself? This question is not only academically challenging but also involves practical issues such as music copyright, originality verification, and the regulation of AI-generated content.

The core challenge of this research lies in determining whether the subtle nuances that distinguish creators' styles still exist and can be quantitatively captured after all music is converted into the uniform and concise YNote format. Specifically, this study aims to address the following key questions:

- Source Classification Challenge: To build an efficient automated system that can accurately classify a YNote music sequence into one of three predefined categories: Native Human Creation, Algorithm Generated, and LLM Generated.

- Information Retention in YNote Format: To investigate whether the simplified YNote notation, designed for machine learning, retains sufficient stylistic information and unique "fingerprints" for effective differentiation and learning by machine learning models during the conversion of complex music into structured text.

- Identification of AI Technological Fingerprints: To study and verify whether different AI generation techniques—namely "Algorithm Generated" representing traditional statistical methods and "LLM Generated" representing deep learning—leave unique and identifiable "technological fingerprints" in their works. Successfully distinguishing between these two is a more profound task than simply differentiating between human and machine creations.

To address these issues, this study reframes the music classification task as a text classification problem in Natural Language Processing (NLP), aiming to verify that through mature text feature extraction techniques, we can uncover key patterns from YNote sequences that distinguish different music sources.

## 3 RESEARCH METHODOLOGY

The goal of this study is to establish an automated classification system capable of accurately distinguishing among three sources of music. The overall process begins with data collection and definition, followed by rigorous data preprocessing and feature engineering, and concludes with model training and evaluation.

### 3.1 Introduction of YNote

To understand the methodology of this study, it's essential to first introduce the data structure at its core: YNote. YNote is a novel, simplified music notation system specifically designed to be machine-learning-friendly, as proposed by Lu et al. [3]. Its primary goal is to overcome the complexity and variable structures of traditional formats like MIDI or ABC Notation, which are not ideal for processing by Large Language Models.

YNote achieves this by representing every musical note as a fixed-length, four-character string. This consistent and concise format effectively translates a complex musical piece into a structured text sequence, making it highly suitable for text-based analysis and machine learning applications. The four-character structure is composed of two key components: Pitch and Duration.

- **Pitch (First 2 Characters):** The pitch is denoted by an English note name (C, D, E, F, G, A, B) followed by a single digit representing the octave. For example, middle C is C4. Notes with a half-step, like C♯ or D♭, are represented by a corresponding lowercase letter (e.g., c). To maintain a consistent format, a rest is represented as 00.

- **Duration (Last 2 Characters):** The duration is also represented by two characters. Basic durations include

01 for a whole note, 02 for a half note, and 04 for a quarter note. The system is designed to handle more complex rhythms, such as dotted notes, triplets, and double-dotted notes, all within the strict two-character limit to ensure structural consistency.

This minimalist design makes YNote both human-readable and, more importantly, highly optimized for machine processing. By converting music into this uniform "language," we can leverage mature Natural Language Processing (NLP) techniques to analyze its structural features, which forms the basis of our classification model. Figure 1 shows an example of YNote notation used to represent Boat on Tai Lake , and Figure 2 provides an overview of the YNote format.

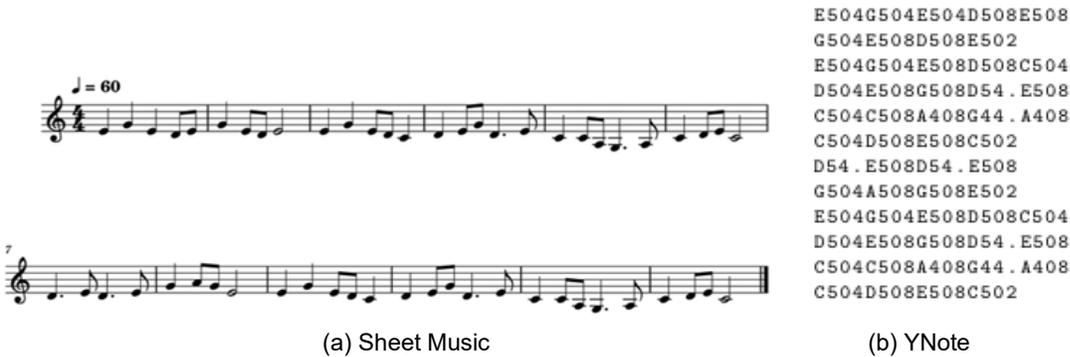

(a) Sheet Music  (b) YNote

Figure 1: Boat on Tai Lake in sheet music and YNote format. Photograph by Lu et al. [3]

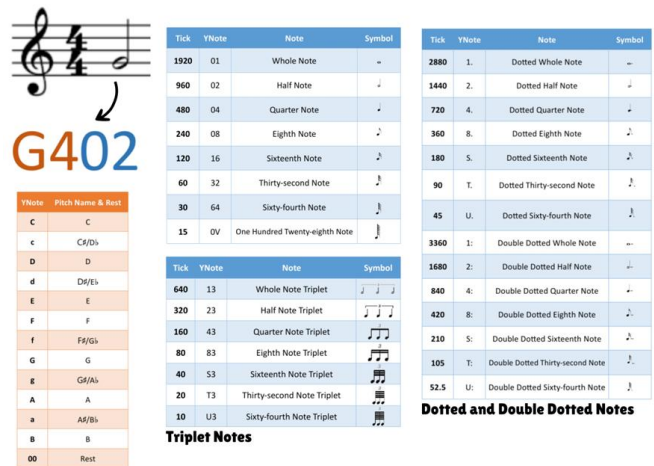

Figure 2: Overview of YNote Format Photograph by Lu et al. [3]

### 3.2 Data Source

The dataset for this study consists of 21,398 songs from three different sources (Native: 669, Algorithm Generated: 18,894, and LLM Generated: 1,835), all uniformly represented in the YNote format. YNote is a simplified music notation system designed for machine learning, where the fixed-length format (4 characters representing a note with pitch and duration) greatly facilitates subsequent automated processing. The significant disparity in the quantity of each category stems from the data collection process. The conversion of native songs from traditional scores to the YNote format currently relies on manual transcription, which inherently limits the scale of the dataset. In contrast, both algorithm-based and LLM-based generation methods can produce a virtually endless supply of musical pieces, resulting in much larger datasets for these categories.

- **Native Songs:** Labeled as class 0, comprising 669 songs. This dataset includes songs of various styles created by human composers, serving as a baseline for human creative style. To ensure data diversity, this category covers multiple genres, including Jiangnan style, Qin style, Gezaixi opera, Japanese Enka, Hakka folk songs, and early Taiwanese pop songs.

- **Algorithm Generated Songs:** Labeled as class 1, comprising 18,894 songs. This dataset is generated by a rule-based and statistical model. It first involves a deep manual analysis of a specific music style (e.g., Jiangnan music) to identify its characteristics in scales, modes, melodic contours, and rhythmic patterns, which are then formalized into rules. Subsequently, the Markov Chain Monte Carlo (MCMC) method is used to calculate the transition probabilities between notes, and the Simulated Annealing Algorithm is employed for melody generation[9].

- **LLM Generated Songs:** Labeled as class 2, comprising 1,835 songs. This dataset is generated by fine-tuning a large language model (e.g., GPT-2). Unlike rule-based algorithms, the LLM autonomously learns the deep structures and patterns of music from a large amount of YNote data, with its output being more focused on data-driven imitation and creation.

### 3.3 Data Preprocessing and Feature Engineering

To convert the raw YNote text into numerical features usable by machine learning models, we performed the following key steps.

#### 3.3.1 YNote Note Sequencing

Before feature extraction, the raw, continuous YNote text string must be converted into a structured sequence of notes. This process is analogous to "tokenization" in natural language processing and is fundamental for machines to understand musical content.

- **Processing Need and Rationale:** The original YNote data is a long, uninterrupted string of characters (e.g., "G402E508C516..."), which machines cannot directly interpret musically. We need to break it down into meaningful basic units "notes." By doing so, we can transform the problem of music analysis into a sequence analysis problem, thereby leveraging mature text processing techniques to explore patterns in music.

- **Processing Method:** Based on the core design of YNote each note (including pitch and duration) is represented by a fixed length of 4 characters—we adopted a deterministic segmentation strategy. Specifically, we iterate through the entire YNote string, cutting it every 4 characters, and treat each resulting segment as an independent note unit. The processed song becomes a sequence of note "tokens" (e.g., ["G402", "E508", "C516", ...]), preparing it for the subsequent TF-IDF feature extraction.

### 3.3.2 TF-IDF Feature Extraction

We chose TF-IDF (Term Frequency-Inverse Document Frequency) as the core feature extraction technique to capture the unique "musical vocabulary" of different music sources.

- **Reason for Selection:** TF-IDF can quantify the importance of a note or melodic fragment (n-gram) to a song. It not only considers the frequency of the fragment in the current song (TF) but also penalizes common fragments that appear universally across all songs (IDF). This allows the model to focus on the unique melodic patterns that best represent the style of a specific category.

- **Key Parameter Settings:**
    1. **n-gram_range=(1, 3):** This is crucial for the model to capture musical patterns. The model considers (1) single notes (1-gram), (2) consecutive pairs of notes (2-grams), and (3) consecutive triplets of notes (3-grams) as features. This enables the model to see not only individual notes but also to learn short melodic fragments and rhythmic combinations.

    2. **max_features=8000:** To control the dimensionality of the feature space and avoid noise, the model selects only the 8000 most frequent n-grams in the dataset as final features.

    3. **min_df=3 & max_df=0.95:** The model ignores rare n-grams that appear in fewer than 3 songs (potential noise) and extremely common n-grams that appear in over 95% of the songs (no discriminative power).

    4. **lowercase=False:** Preserves the original case of YNote. This is vital in YNote, as lowercase letters are often used to denote sharps or flats, and preserving the case retains the complete musical information.

### 3.3.3 Dataset Splitting

To objectively evaluate the model's performance, the dataset was split into training, validation, and test sets.

- **Split Ratios:**
    1. **Training Set:** 65%
    2. **Validation Set:** 15%
    3. **Test Set:** 20%

- **Stratified Sampling (stratify=y):** This is an important strategy to ensure the reliability of the evaluation. Through stratified sampling, the script ensures that the proportions of the three music categories (Native, Algorithm, LLM) in the split training, validation, and test sets are consistent with the original dataset, avoiding evaluation bias due to uneven random sampling.

*3.3.4 SMOTE for Data Imbalance*

The original dataset has a significant disparity in the number of songs per category (Native: 669, Algorithm: 18,894, LLM: 1,835). To prevent the model from being biased towards the majority class, we applied SMOTE (Synthetic Minority Over-sampling Technique) only to the training set.

- **Processing Method:** SMOTE synthesizes new samples by performing linear interpolation between minority class samples and their nearest neighbors in the feature space. In this study, we oversampled the "Native Songs" and "LLM Generated Songs," which have fewer samples, to match the number of "Algorithm Generated Songs," ultimately achieving a balanced 1:1:1 ratio among the three classes in the training set.

- **Key Parameter Settings:**
    1. **random_state=42:** Ensures that the oversampling results are consistent across runs, making the experiment reproducible.
    2. **k_neighbors:** The value of parameter k is dynamically set to min(5, min_samples - 1). This is a robust design that ensures k does not exceed the total number of minority samples minus one, preventing errors in the SMOTE algorithm when the minority class has very few samples.

### 3.4 Classification Model

This study selected Logistic Regression as the final classifier.
- **Reason for Selection:**

    **Efficiency and Interpretability:** Logistic Regression models are fast to train and consume low computational resources, making them well-suited for handling the sparse, high-dimensional features generated by TF-IDF. More importantly, its model coefficients can directly reflect the influence of each feature (i.e., musical n-gram) on the classification decision, providing excellent interpretability.

- **Model Parameter Settings:**
    1. **multi_class='ovr':** Adopts the "One-vs-Rest" strategy to handle the three-class problem. The model trains a separate binary classifier for each class, which is responsible for distinguishing the "current class" from "all other classes."

    2. **class_weight='balanced':** As an auxiliary data balancing strategy, this parameter automatically assigns higher weights to classes with fewer samples, increasing the penalty for misclassifying minority classes during the loss calculation in model training.

    3. **solver='liblinear':** Chooses liblinear as the optimization algorithm, which performs robustly and efficiently on small to medium-sized datasets and high-dimensional sparse data.

    4. **max_iter=2000:** Sets the maximum number of iterations for the solver to converge to 2000, ensuring the model has sufficient time to find the optimal solution on complex data.

    5. **random_state=42:** Also for ensuring the reproducibility of the experimental results.

## 4 RESEARCH RESULTS

The model was finally evaluated on an independent test set, where it achieved outstanding performance. This section presents detailed evaluation results from multiple perspectives.

### 4.1 Overall Performance Metrics

The model demonstrated excellent performance across several key metrics, proving its robustness and high precision.

- **Three-Class Accuracy:** 98.25%

- **5-fold CV Accuracy:** 0.9857 ± 0.0013

- **ROC-AUC Scores:**
    1. **Micro-average ROC-AUC:** 0.9988
    2. **Macro-average ROC-AUC:** 0.9934

- **Per-Class ROC-AUC Scores:**
    1. **Native Songs:** 0.9904
    2. **Algorithm Generated:** 0.9927
    3. **LLM Generated:** 0.9971

An accuracy of 98.25% and stable performance with a low standard deviation in cross-validation indicate that the model has good generalization ability. The ROC-AUC scores, being close to 1, show that the model has excellent discriminative power across all classes.

## 4.2 Classification Report and Confusion Matrix

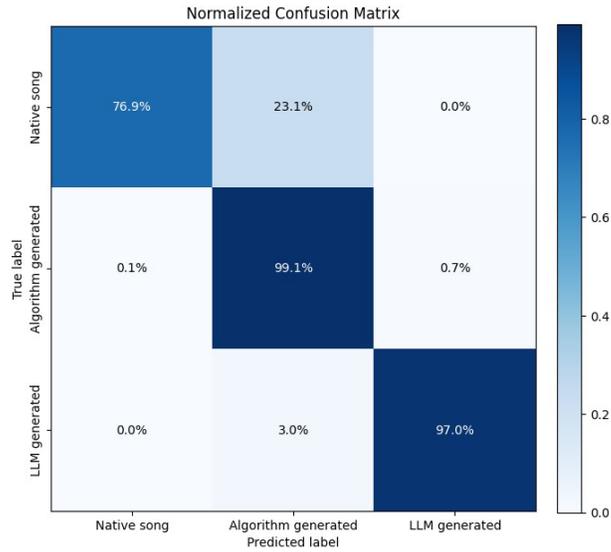

Figure 2: Normalized Confusion Matrix for the Logistic Regression Model on the Test Set

Table 1: Table of Classification Report on the Test Set

| Class | Precision(%) | Recall(%) | F1-Score | Support |
|---|---|---|---|---|
| Native Songs | 0.95 | 0.77 | 0.85 | 134 |
| Algorithm Generated | 0.99 | 0.99 | 0.99 | 3779 |
| LLM Generated | 0.93 | 0.97 | 0.95 | 367 |
| Weighted average / Total | 0.98 | 0.98 | 0.98 | 4280 |

Analyze Table 1 and Figure 2 in detail according to the following description:

- Native Songs: The model's recall for this class is 0.78, the lowest among the three, meaning that 22% of native songs were missed (primarily misclassified as algorithm-generated). However, its precision is high at 0.95, indicating that when the model predicts a song is native, it is correct 95% of the time. This suggests that native songs have a more diverse style, and the structural features of some works may overlap with those of algorithm-generated music.

- Algorithm Generated Songs: All metrics for this class reached 0.99, showing near-perfect performance. This indicates that music generated by MCMC and statistical rules has very distinct, consistent, and stable patterns that are easily captured and identified by the model.

- LLM Generated Songs: The model also performed excellently in this category, with an F1-Score of 0.95. A high recall of 0.97 shows that the model can successfully identify most LLM-generated songs. A small amount of confusion (11 songs) was directed towards "Algorithm Generated Songs," possibly because both AI generation techniques share commonalities at some basic musical syntax levels.

### 4.3 Important Feature Analysis

By examining the coefficients of the logistic regression model, we can identify the "musical vocabulary" (n-grams) that contribute most to the classification of each category. A positive coefficient indicates that the feature is a strong predictor for that class, while a negative coefficient indicates the opposite.

Table 2: Top 10 Important Features for Each Music Source Category and Their Logistic Regression Coefficients

| rank | Native Songs | | Algorithm Generated | | LLM Generated | |
|---|---|---|---|---|---|---|
| | n-gram | Coefficient | n-gram | Coefficient | n-gram | Coefficient |
| 1 | 0002 | 8.8503 | 0004 | -9.6427 | A502 D68. | 6.8798 |
| 2 | 0004 | 7.8124 | 0002 | -8.1328 | E508 | 6.4754 |
| 3 | 0008 | 7.6662 | E408 | -7.4674 | C616 | 5.2655 |
| 4 | G508 G516 | 7.2089 | 0008 | -7.1655 | G402 E508 | 4.9406 |
| 5 | G516 G516 | 6.8737 | C608 C608 | -6.8140 | G502 G516 | 4.9283 |
| 6 | A508 A516 | 6.6125 | E608 E608 | -6.7450 | C504 D504 | 4.6730 |
| 7 | E608 E608 | 6.4250 | A508 A508 | -6.4580 | E616 | 4.6234 |
| 8 | C608 C608 | 6.3628 | C508 C508 | -6.2580 | G516 | 4.6083 |
| 9 | C508 C508 | 5.9568 | F508 | -6.2232 | D616 E616 | 4.5530 |
| 10 | E408 | 5.9130 | E416 | -6.1164 | D502 G508 | 4.4111 |

According to Table 2 the feature analysis reveals that features representing rests, such as 0002, 0004, and 0008, play a key role in distinguishing between native and algorithm-generated music. Native songs tend to use rests of these durations more frequently, whereas algorithm-generated music significantly avoids them (as shown by large negative coefficients). LLM-generated music exhibits some unique 2-gram combinations, like A502 D68. and G402 E508, which are likely specific patterns learned from the data.

### 5 CONCLUSION

This study proposes and validates an efficient automated framework for music source classification. By innovatively combining the YNote music format, designed for machine learning, with the TF-IDF feature extraction technique from natural language processing, we not only achieved a classification accuracy of up to 98.25% but also opened a new and effective path for the content analysis of symbolic music.

The success and innovative value of this research are reflected in the following aspects:

1. **Methodological Innovation and Effectiveness:** This study is a successful example of applying the mature n-gram and TF-IDF models from the NLP domain to the classification of YNote symbolic music. The results strongly demonstrate that a musical sequence can be treated as a special "language," whose intrinsic style and structural patterns can be efficiently captured and quantified. The 98.25% accuracy is an objective validation of this methodology's effectiveness.

2. **Deep Validation of Model Learning Ability:** The success of this model is not just reflected in its high accuracy. The classification report shows that the model achieved high precision and recall across all categories. This is crucial because it proves that the model did not achieve high accuracy through guessing or exploiting data biases, but rather by genuinely learning the intrinsic features of each music category. High precision means the model's predictions are reliable; high recall means the model can comprehensively identify samples of a specific class. The combination of both indicates that our feature engineering and model selection were successful, and the model has indeed captured the key "fingerprints" that distinguish different music sources.

3. **Validation of YNote Format's Potential:** The excellent performance of this model, in turn, validates the great potential of YNote as a music representation method designed for machine learning. Its uniformity, conciseness, and structured nature make it an ideal bridge connecting musical art and data science, greatly lowering the barrier to music data analysis.

4. **Deep Insights into AI-Generated Music:** The contribution of this research extends beyond distinguishing between "human-created" and "AI-generated" music. More valuable is the model's ability to accurately differentiate between two distinct AI generation paradigms—rule-based algorithms and data-driven LLMs. This reveals that different AI technologies leave their own unique, identifiable "technological fingerprints" in musical creation, providing a powerful tool for the future tracing, evaluation, and auditing of AI-generated content.

In summary, this study not only completed a specific classification task but, more importantly, proposed and validated a complete, efficient, and highly interpretable framework for music style analysis. The success of this framework lays a solid foundation for future research in areas such as music style recognition, composer intent analysis, and music copyright tracking, demonstrating the broad application prospects of data science in the field of musical arts.

## 6 LIMITATIONS AND FUTURE OUTLOOK

Although this study has successfully built a high-precision music source classification model, there are still some limitations, which also point the way for future research.

### 6.1 Limitations

1. **Singularity of Data Format:** This research is based entirely on the YNote format, which is designed for machine learning. While the simplicity of YNote was key to the success of this study, it also discards many subtle details found in traditional scores, such as dynamics, timbre, and articulation. Therefore, the findings of this model may not be directly generalized to the analysis of richer music formats like MIDI or raw audio.

2. **Specificity of the Dataset:** The "Algorithm Generated Songs" dataset used in this study was primarily generated based on rules derived from a specific style of Jiangnan music. This may have led the model to learn an "algorithmic fingerprint" that is biased towards this style. The model's generalization ability, its performance in identifying algorithm-generated music from other cultural backgrounds or styles needs further validation.

3. **Limitations of Feature Extraction Method:** The TF-IDF and n-gram models used in this study mainly capture local note sequence patterns (combinations of 1 to 3 notes). While this method has proven to be very effective, it may struggle to capture more long-term structures, hierarchical relationships, or contrapuntal textures in music.

4. **Rapid Iteration of AI Technology:** The LLM songs in this study were generated by fine-tuning a GPT-2 level model. With the rapid development of generative AI technology, music generated by newer, more powerful models may be closer to human creations, and their "technological fingerprints" may become more subtle, posing an ongoing challenge to the identification capabilities of the current model.

**6.2 Future Outlook**

1. **Exploring More Advanced Sequence Models:** In the future, more advanced natural language processing models, such as Recurrent Neural Networks (RNNs), Long Short-Term Memory (LSTM) networks, or even Transformer architectures, could be applied to YNote sequence analysis. These models are better at capturing long-term dependencies in data and may be able to identify complex musical structures that TF-IDF cannot.

2. **Expanding and Diversifying the Dataset:** A key focus for future work is to build a dataset that includes a more diverse range of cultures, genres, and historical periods. This will help train a more generalizable classifier and verify whether the "technological fingerprints" discovered in this study are universally applicable across different styles.

3. **Conducting Feature Interpretability Analysis:** Future research could delve deeper into analyzing the most important TF-IDF features learned by the model. By identifying which "musical words" are characteristic of humans, algorithms, or LLMs, we can not only enhance the model's credibility but also potentially gain profound musicological insights, revealing the essential differences in micro-melodic patterns among different creators.

4. **Cross-Modal Integrated Research:** In the long term, we could explore combining the symbolic music analysis framework of this study with audio-based spectral analysis. Creating a cross-modal model that can process both YNote symbolic data and audio features could capture source information from different dimensions, leading to more robust and comprehensive detection of AI-generated content.

# Authors' background

| Your Name | Title* | Research Field | Personal website |
|---|---|---|---|
| Cheng-Yang Tsai | Undergraduate Student | EDA, Algorithm, AI music | N/A |
| Tzu-Wei Huang | Undergraduate Student | EDA, Algorithm, AI music | N/A |
| Shao-Yu Wei | Undergraduate Student | EDA, Algorithm, AI music | N/A |
| Guan-Wei Chen | Undergraduate Student | EDA, Algorithm, AI music | N/A |
| Hung-Ying Chu | Undergraduate Student | EDA, Algorithm, AI music | N/A |
| Yu-Cheng Lin | Assistant Professor | EDA, Algorithm, AI music | https://www.cse.yzu.edu.tw/en/people/professor?name=Yu-Cheng%20Lin |